\documentclass[journal]{IEEEtran}

\usepackage{fontspec}
\usepackage{academicons}

\usepackage{amsmath,amssymb,amsfonts}   
\usepackage{graphicx}                   
\usepackage{booktabs}                   
\usepackage{multirow}                   
\usepackage{cite}                       
\usepackage{capt-of}
\usepackage[ruled,linesnumbered]{algorithm2e} 
\usepackage{xcolor} 
\usepackage[hidelinks]{hyperref} 

\newcommand{\R}{\mathbb{R}}             
\newcommand{\orcidicon}[1]{\href{https://orcid.org/#1}{\raisebox{0.3\height}{\scalebox{0.82}{\textcolor[HTML]{A6CE39}{\aiOrcid}}}}}

\graphicspath{{figures/}}

\begin{document}

\bstctlcite{BSTcontrol}

\title{UniPoint: Unified Point-Level Sensor Fusion for Humanoid Locomotion Across Challenging Terrains}

\author{Sicen~Li\orcidicon{0000-0002-2049-8921}\textsuperscript{1},
        Zhen~Chu\orcidicon{0009-0000-6556-3427}\textsuperscript{1},
        Chao~Li\orcidicon{0009-0005-7199-1985}\textsuperscript{2},
        Qiuguo~Zhu\orcidicon{0000-0002-4965-5126}\textsuperscript{1},
        and~Jun~Wu\orcidicon{0000-0002-1388-7451}\textsuperscript{1,3,*}%
\thanks{\textsuperscript{1}College of Control Science and Engineering,
Zhejiang University, Hangzhou 310027, China (e-mail:
0625416@zju.edu.cn; 12560199@zju.edu.cn; qgzhu@zju.edu.cn;
junwuapc@zju.edu.cn). 

\textsuperscript{2}Yunshenchu Technology
Co., Ltd., Hangzhou, China (e-mail: lichao@deeprobotics.cn).

\textsuperscript{3}Zhejiang
Key Laboratory of Additive Manufacturing Technology and Equipment,
Hangzhou, China. 

\textsuperscript{*}Corresponding author: Jun Wu.}%
\thanks{Video: \url{https://youtu.be/Rd9YyfOxvmY}}%
\thanks{This work has been submitted to the IEEE for possible publication. Copyright may be transferred without notice, after which this version may no longer be accessible.}}

\markboth{IEEE Robotics and Automation Letters. Preprint version.}%
{Li \MakeLowercase{\itshape et al.}: UniPoint: Unified Point-Level
Sensor Fusion for Humanoid Locomotion}

\maketitle

\begin{abstract}
Open-world deployment requires humanoid robots to cross highly
heterogeneous terrain safely, with perception that simultaneously
provides wide coverage, local accuracy, and redundancy against sensor
failure. Existing approaches struggle to satisfy all three: one
forward depth camera or nearby height sampling covers too little;
odometry-corrected elevation maps drift under aggressive motion
and miss thin vertical
structures; image-level encoding costs grow with camera count. We
present UniPoint, a humanoid whole-body locomotion framework built on
multi-source point-level sensor fusion. Measurements from a
$360^{\circ}$ light detection and ranging (LiDAR) sensor and two
depth cameras are early-fused into one base-frame point set. Voxelization
resamples it to a fixed number of tokens encoded by linear
self-attention and proprioception-queried cross-attention, decoupling
forward cost from sensor count. The point set retains standing thin barriers;
a single-modality failure removes only part of the tokens, so the
policy degrades gracefully. A single training run with terrain-aware
rewards, perception-degradation injection, and domain randomization
produces one policy for all eight terrain types, deployed on an
onboard RK3588 without fine-tuning. On a DR02 humanoid, 20 trials at each of
nine real-world settings over seven terrain types validate the policy
on 70-cm-high platforms, 100-cm gaps, thin barriers, and sparse or
narrow footholds; it also generalizes zero-shot outdoors.
\end{abstract}


\IEEEpeerreviewmaketitle

\begin{figure}[!t]
\centering
\includegraphics[width=\columnwidth]{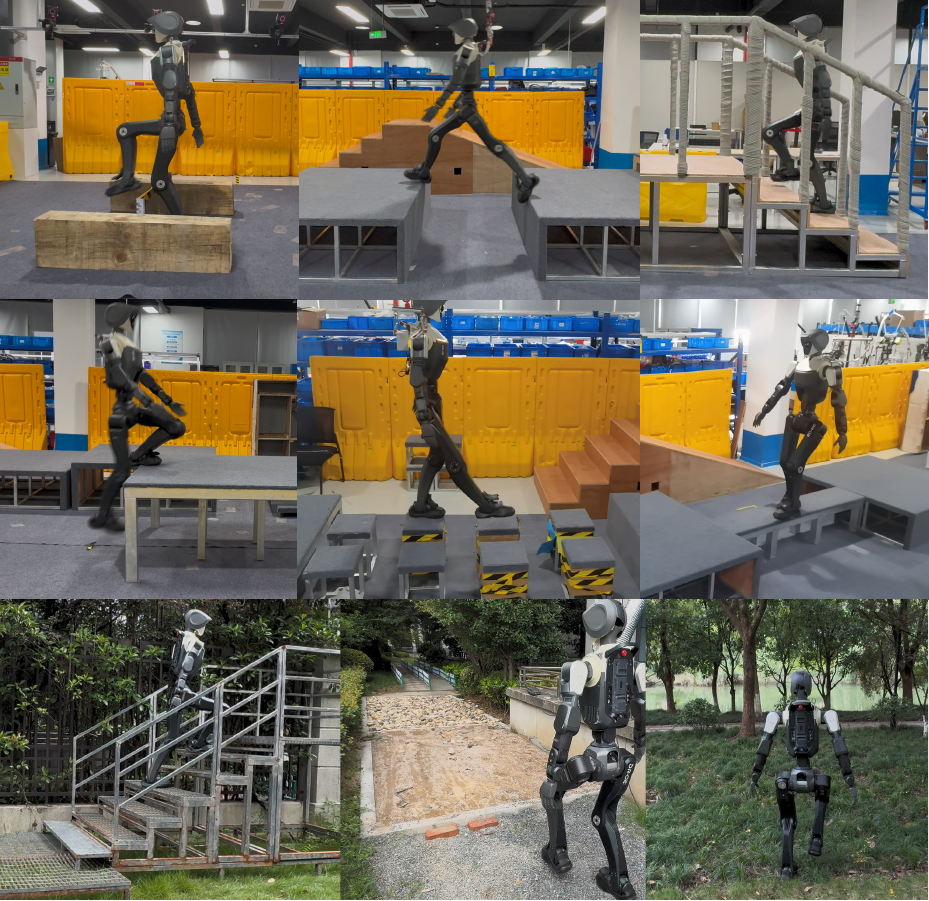}
\caption{DR02 humanoid traverses challenging terrains with
a single policy in real-world tests and generalizes zero-shot to three
outdoor scenes. Top row, left to right: 45-cm thin barrier, 100-cm gap,
20-cm stairs. Middle row: 70-cm-high platform, stepping stones, balance
beam. Bottom row (outdoor): industrial steel-grating stairs, gravel,
grass.}
\label{fig:teaser}
\end{figure}

\section{Introduction}
\label{sec:intro}

Humanoid robots are entering open-world applications from substation
inspection to industrial maintenance, where whole-body locomotion must
cross highly heterogeneous terrain: stairs, slopes, discrete
obstacles, high platforms, gaps, stepping stones, balance beams, and
thin barriers (Fig.~\ref{fig:teaser}). Safe traversal demands
footholds inside shrinking support regions, coordinated whole-body
motion, and locomotion that survives unreliable perception.

Proprioception-only policies~\cite{lee2020quadrupedal,long2024hybrid}
are cheap and immune to perception failure but cannot anticipate
terrain beyond the feet, a structural limit on sparse footholds.
Vision-based routes gain foresight from depth images or elevation
maps but face four structural problems:
1)~a single forward camera~\cite{luo2024pie,sun2026dpl} or height
sampling around the robot~\cite{long2025learning} covers too little;
2)~elevation maps corrected by odometry drift under aggressive motion
and lose thin vertical structures (Section~\ref{sec:real});
3)~exteroception and proprioception fail separately, leaving a
persistent sensor-trust tradeoff~\cite{miki2022learning}; and
4)~image-level encoding costs grow with camera
count~\cite{zhang2026rpl}.

We present UniPoint, a humanoid whole-body locomotion framework built
on point-level fusion: it concatenates the measurements of a
$360^{\circ}$ light
detection and ranging (LiDAR) sensor and front and rear depth cameras
into one base-frame point set, voxelizes
it into a fixed number of tokens, and encodes the tokens with
linear self-attention; cross-attention fuses them with the
proprioceptive history, without building any image or elevation map.
The fused points give near-omnidirectional coverage, retain standing
thin barriers, and form odometry-free short-term memory. The token
budget caps the forward cost, and a single-modality failure
deletes only part of the tokens, so the policy stays functional.

Our contributions are as follows: \textbf{1) Point-level multi-source
early fusion.} Any range sensor is just a source of 3D points: we
early-fuse the points of a $360^{\circ}$ LiDAR and two depth
cameras into a fixed number of voxel tokens, with the forward cost
decoupled from sensor
count; the forward pass takes 1.5~ms on an RK3588, under one-fifth
that of a depth-image convolutional neural network (CNN). The 3D point set
retains standing thin barriers; height sampling largely fails on them,
18\% against UniPoint's 99.8\% in simulation (Section~\ref{sec:sim}).
\textbf{2)~Unified training with deployment-grade robustness.} Two
terrain-aware rewards, a foot-sole support-integrity scan and
slope-aligned velocity decomposition, turn the safe-foothold
requirement into dense reward signals (Section~\ref{sec:reward});
removing the foot-sole scan drops stepping-stone success from 79\% to
68\% and stair success from 99\% to 92\% in simulation
(Section~\ref{sec:sim}). With perception-degradation injection and
domain randomization, a single unified training run produces one policy
that covers all eight terrain types and transfers to the real robot
without fine-tuning (Sections~\ref{sec:training} and~\ref{sec:real}).
\textbf{3)~System-level validation on a full-size, low-compute
platform.} On the DR02 humanoid (1.75~m, 75~kg, RK3588 onboard), 20
trials at each of nine settings over seven terrain types show the
single policy crossing 20-cm stairs, 70-cm-high platforms, and 100-cm
gaps, exceeding the 45-cm platforms and 80-cm gaps reported on the same
robot~\cite{yu2026ssr}. The blind, proprioception-only baseline
retains only stairs
and slopes; the elevation-map baseline fails completely on thin
barriers and degrades sharply on harder terrain. With the LiDAR fully
occluded, UniPoint still crosses a 40-cm-high platform and stairs,
and generalizes zero-shot outdoors (Section~\ref{sec:real}).

\section{Related Work}
\label{sec:related}

\subsection{Learning-Based Humanoid Locomotion}
\label{sec:rw_locomotion}

Building on massively parallel reinforcement
learning~\cite{rudin2021learning} and privileged
learning~\cite{lee2020quadrupedal}, humanoid work has branched
according to perception input: PIM~\cite{long2025learning} samples heights around
the robot; DPL~\cite{sun2026dpl} uses one depth camera;
RPL~\cite{zhang2026rpl} distills multiple cameras for bidirectional
walking; Song~et~al.~\cite{song2025gait} reconstruct under-base
terrain for backward walking, but only nearby; SSR~\cite{yu2026ssr}
reaches reliable footholds and 1.3~km outdoors in one stage;
DWL~\cite{gu2024advancing} and HPL~\cite{zhuang2024humanoid} cross
challenging terrain with a denoising world model and an end-to-end
vision policy, respectively. UniPoint fuses a $360^{\circ}$ LiDAR with two depth cameras at the point level,
covering eight terrain types for low-compute onboard deployment.

\subsection{Perceptive Locomotion and Terrain Representation}
\label{sec:perception}

Privileged learning trains a teacher on height samples around the
robot and distills a proprioception-only
student~\cite{lee2020quadrupedal}; Miki~et~al.~\cite{miki2022learning}
add belief-gated fallback to proprioception on noisy samples;
PIE~\cite{luo2024pie} estimates terrain explicitly and implicitly;
START~\cite{yu2026start} and DPL~\cite{sun2026dpl} reconstruct
elevation maps to ease sparse-foothold exploration;
GLAD~\cite{fu2026glad} splits terrain encoding into route and foothold
branches; TAGA~\cite{li2026taga} learns an active gaze (1.2-m
real-world gap); Bank~et~al.~\cite{bank2026hybrid} fuse LiDAR and
depth but still through elevation maps.

Point-cloud interfaces are emerging: DreamWaQ++~\cite{nahrendra2026dreamwaq}
pools PointNet~\cite{qi2017pointnet} features into a context vector;
Wang~et~al.~\cite{wang2025endtoend} process single-LiDAR sequences end
to end; Omni-Perception~\cite{wang2025omni} hierarchically encodes
a $360^{\circ}$ LiDAR; Gallant~\cite{ben2025gallant} voxelizes LiDAR for
overhead-constrained terrain; AME~\cite{he2025attention} applies
proprioception-conditioned attention to steppable map regions; and
CReF~\cite{hao2026cref} queries depth features with proprioception by
cross-attention. UniPoint early-fuses the points of every sensor into
one voxel grid and aggregates them by proprioception-queried
cross-attention, without any mapping or odometry stack, at a cost
independent of sensor count
(Sections~\ref{sec:fusion} and~\ref{sec:real}).

\subsection{Safe Foot Placement and Reward Design}
\label{sec:rw_foothold}

BeamDojo~\cite{wang2025beamdojo} designs a sampling-based foothold
reward for polygonal feet and balances its sparse signal with a double
critic, crossing 20-cm stepping stones at 45-cm spacing zero-shot on a
Unitree G1; MARG~\cite{dong2025marg} crosses 65-cm gaps and an 18-cm
balance beam with foot-related rewards and a single-LiDAR elevation
map; waypoint velocity inner products~\cite{cheng2024extreme} prevent
detours, and foot-edge
penalties~\cite{cheng2024extreme,yu2026start} prevent edge landings;
concurrent work~\cite{jiang2026omnidirectional} walks stairs
omnidirectionally with a dense unsafe-stepping penalty and
sparse-LiDAR elevation maps. Our foot-sole scan shares this lineage:
it judges overhang and penetration bidirectionally with downward
rays, densifies the violating-point signal within stance, and couples
with slope-aligned velocity decomposition in one eight-terrain
curriculum (Section~\ref{sec:reward}).

\section{Method}
\label{sec:method}

\begin{figure*}[!t]
\centering
\includegraphics[width=\textwidth]{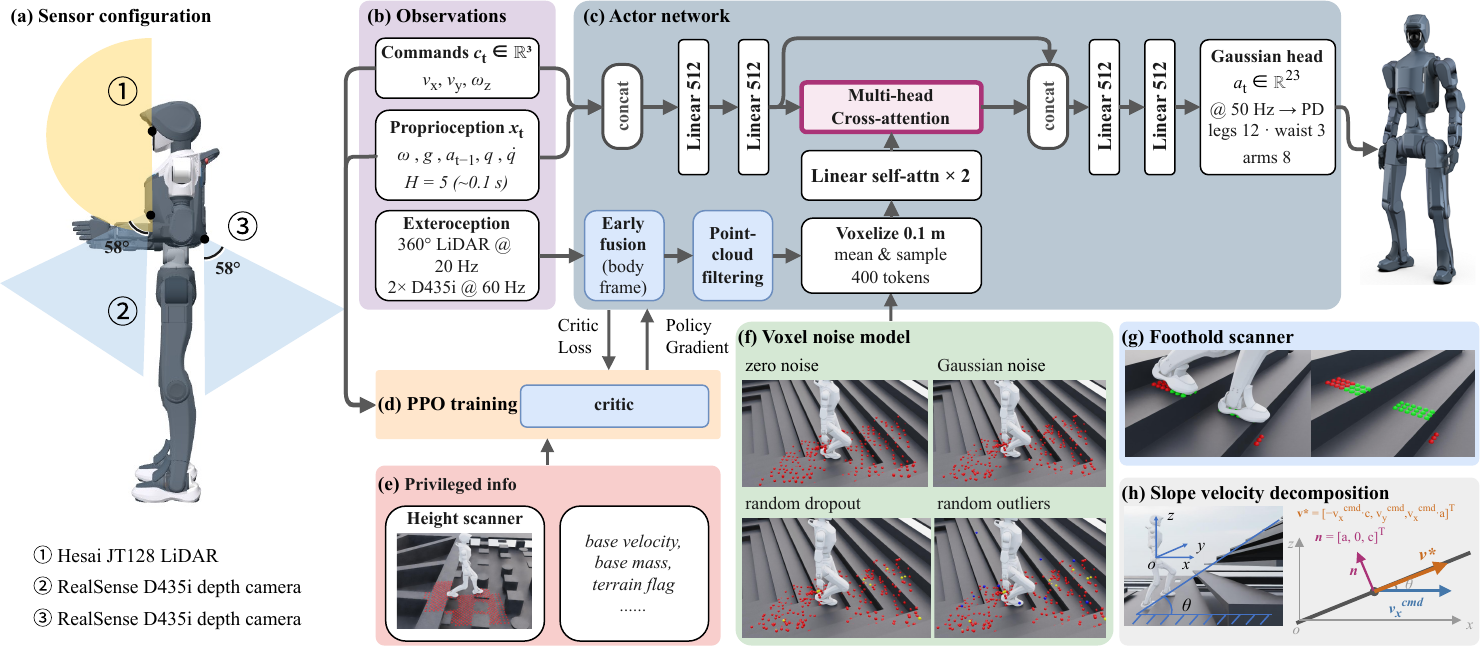}
\caption{Framework overview. (a)~Sensor configuration: JT128
$360^{\circ}$ LiDAR and front and rear D435i depth cameras.
(b)~Observations: commands, stacked proprioception, and multi-source
exteroception. (c)~Onboard actor network, the policy of
Sections~\ref{sec:overview} and~\ref{sec:fusion}. (d)~PPO training:
the critic loss from (e) updates the actor in (c) by policy
gradient. (e)~Privileged information: the asymmetric critic
additionally receives a height scan, base velocity, base mass, and
terrain flag on top of proprioception and commands.
(f)~Voxel noise model: training-time cloud degradation, red
original, yellow dropped, blue outlier points. (g)~Foothold
scanner: foot-sole support-integrity scan; green samples lie
inside the support region, red ones overhang. (h)~Slope-aligned
velocity decomposition, base upright. Panels (d)--(h) exist only
in training; deployment uses only (a)--(c).}
\label{fig:framework}
\end{figure*}

\subsection{Overview}
\label{sec:overview}

We formulate whole-body locomotion as a
partially observable Markov decision process: the policy
$\pi_{\theta}$ outputs joint position targets $a_t \in \R^{23}$ (12
leg, 3 waist, and 8 arm joints) at 50~Hz, tracked by a low-level
proportional--derivative (PD) controller. Its observation
combines the proprioceptive history, the voxel token set
$\mathcal{P}_t$ of Section~\ref{sec:fusion}, and the command
$c_t \in \R^{3}$. The proprioceptive vector $x_t \in \R^{75}$ collects
base angular velocity, projected gravity, the previous action, and joint
positions and velocities; stacking the last $H{=}5$
frames, about 0.1~s, eases partial observability, and training
samples longitudinal commands from $-1.0$ to $1.2$~m/s, lateral
$\pm 0.55$~m/s, and yaw rate $\pm 1.2$~rad/s. Training maximizes the
discounted return $J(\theta) = \mathbb{E}_{\tau \sim
\pi_\theta}[\sum_{t} \gamma^{t} r_t]$, $\gamma = 0.99$, with
proximal policy optimization (PPO)~\cite{schulman2017ppo}.

Training is single-stage and end-to-end, with no pretraining,
distillation, hierarchical planning, or gait constraint: no gait
clock, phase signal, or contact schedule enters the policy, and gait
rhythm emerges from reward shaping and the terrain curriculum. The
asymmetric critic additionally receives privileged information,
height-scan grids around
the base, per-link masses, true action delays, a terrain-type flag,
and noise-free kinematics, for value estimation only.

Our central setting is \textit{unified training}: one policy trains
once on a combined curriculum over all eight terrain types, with
no terrain switch, expert gating, or per-terrain fine-tuning at
deployment. Thin barriers are plates standing upright on the
ground, with thickness far
below the foot length; they are nearly invisible
to two-dimensional height representations (Section~\ref{sec:sim}).

\subsection{Multi-Source Point-Level Perception Fusion}
\label{sec:fusion}

Safe footholds demand wide coverage and local accuracy at once:
depth cameras are dense but narrow and occlusion-prone, and a LiDAR
covers the surroundings widely at low point density. We fuse all
measurements into one base-frame point set before
any encoding, so the representation stays independent of
sensor identity (Fig.~\ref{fig:framework}(a)). The sensors are a
head-mounted Hesai JT128 hyper-hemispherical LiDAR,
$360^{\circ} \times 189^{\circ}$ at about 20~Hz, and front and rear
torso-mounted RealSense D435i depth cameras, about
$87^{\circ} \times 58^{\circ}$ at 60~Hz.

Three steps normalize the points into a fixed-size token set. We
drop points inside the torso bounding box; swinging limbs can still
extend beyond it and appear in the cloud (Section~\ref{sec:real}). The
rest are quantized onto a 0.1-m voxel grid in the base frame (the robot's
root link), covering
1.3~m ahead, 1.1~m behind, $\pm$0.7~m sideways, 1.4~m below, and
0.2~m above the base, each occupied voxel replaced by its
interior-point mean. We then resample a fixed $K{=}80$ points
uniformly at random, without replacement, from the occupied voxels; the
same random rule serves training and inference, and any shortfall is
padded with masked invalid tokens. Together with the last $G{=}5$ frames (independent of proprioceptive
$H$), the policy input is
$\mathcal{P}_t = \{ p_i^{(j)} \in \R^{3} \mid i = 1,\dots,K,\;
j = t{-}G{+}1,\dots,t \}$, that is, $80 \times 5 = 400$ voxel tokens
per step, each point's coordinates augmented with a frame
position encoding.

The fixed token budget doubles as density normalization and an
information bottleneck: the token count and structure stay constant
whatever the source, and the complementary 20-Hz LiDAR and 60-Hz
cameras make the frame stack a
short-term terrain memory. We measure and voxelize each frame in
the base frame at its own capture time, with no cross-frame
re-registration. The five frame grids are mutually offset by the
base displacement over the stacking window, while the frame
position encoding tells the policy each token's frame age, so it
implicitly compensates for this offset. Short-term memory
therefore
needs no odometry; the LiDAR's rotational-sweep distortion also
goes uncompensated, absorbed by policy robustness and multi-source
redundancy. A single-modality failure
deletes only part of the tokens, and the policy falls back
to the remaining modalities and proprioception
(Sections~\ref{sec:sim} and~\ref{sec:real}).

The onboard network (Fig.~\ref{fig:framework}(c)) has two encoding
paths and one fusion module. The point path first concatenates each
point's coordinates with its frame position encoding into
$[Batch, 400, 4]$ tokens and then lifts them point-wise to
$[Batch, 400, 32]$ features through two linear self-attention
layers~\cite{katharopoulos2020linear}.

The proprioception path passes the $H$-frame history
and the command through two 512-dimensional linear layers into a
state vector $s_t$. Fusion is proprioception-queried
cross-attention: 16 softmax heads of dimension 32, queries
projected from $s_t$, keys from projected point features, and
values the point features themselves. The head contexts and $s_t$
feed two further 512-dimensional linear layers, and a Gaussian head outputs the mean
and log standard deviation of the 23-dimensional action
distribution, using the mean at
deployment. Actions are residuals around the default stance, truncated
at the joint limits. Using $s_t$ as the queries lets the policy
condition on its own state and dynamically choose which spatial
points to attend to; the near-omnidirectional coverage also lets the
robot sense obstacles behind it without turning
(Section~\ref{sec:real}). The kernel approximation lowers the
multiply-adds of the attention blocks from $O(N^2 d)$ to
$O(N d^2)$, a theoretical saving of about $25\times$ at $N{=}400$,
$d{=}16$ for the attention kernels alone (about $5\times$ end to
end; Section~\ref{sec:setup}), at matched accuracy
(Section~\ref{sec:sim}).

\subsection{Terrain-Aware Reward Design}
\label{sec:reward}

The reward is a weighted sum over the terms of
Table~\ref{tab:reward}; its two terrain quantities, the foot-sole
scan and the slope plane fit, are privileged simulation queries
used only in reward calculation, and
the deployed policy reads no height scan.

\begin{table}[!t]
\caption{Reward terms and weights}
\label{tab:reward}
\centering
\scriptsize
\setlength{\tabcolsep}{1.5pt}
\begin{tabular}{@{}llc@{}}
\toprule
Reward term & Form & Weight\\
\midrule
Linear velocity tracking & $\exp(-\|\boldsymbol{e}_{t,xy}\|^2/\sigma_v^2)$ & 0.6\\
Vertical velocity tracking & $\exp(-e_z^2/\sigma_v^2)$, slope-aligned & 0.3\\
Yaw-rate tracking & $\exp(-(\omega_z - \omega^{*})^2/\sigma_\omega^2)$ & 0.5\\
\midrule
Foot-sole scan & (\ref{eq:sole}), stance-gated & 0.5\\
\midrule
Base orientation & $1{-}\exp(-\|\boldsymbol{g}{-}\hat{\boldsymbol{g}}\|^2/\sigma_g^2)$ & $-1.0$\\
Base height & $1{-}\exp(-[(h^{*}{-}h)^{+}]^2/\sigma_h^2)$ & $-1.0$\\
Actuator effort & $-\sum_{i}0.055\tilde{\tau}_i^2$ & ---\\
Joint velocity & $-\sum_{i}0.1\tilde{q}_i^2$ & ---\\
Action rate & $\|a_t{-}a_{t-1}\|^2$ & $-0.04$\\
Action smooth & $\|a_t{-}2a_{t-1}{+}a_{t-2}\|^2$ & $-0.02$\\
Limit violations & $\sum_{i}\mathbf{1}[\mathrm{viol}_i]$ & $-0.2$\\
Foot slip & $\sum_{f\in\mathcal{F}_{\mathrm{st}}}\|\boldsymbol{v}_f\|^2 / v_{\mathrm{co}}^2$ & $-0.2$\\
Foot impact & $\sum_{f\in\mathcal{F}_{\mathrm{st}}}(a_f - 60)^{+}$ & $-0.05$\\
Illegal contact & $\sum_{b\in\mathcal{B}_{\mathrm{ill}}}\mathbf{1}[\mathrm{contact}_b]$ & $-0.5$\\
\bottomrule
\end{tabular}
\end{table}

\textbf{Foot-sole support-integrity scan.} On stairs, high
platforms, stepping stones, balance beams, and gaps, landing inside
the support region is necessary for safety, yet falling itself is a
sparse and delayed signal. The foot-sole scan turns this requirement
into a dense, stance-gated penalty (Fig.~\ref{fig:framework}(g)). A
grid of $M{=}21$ points, $7 \times 3$ at about 0.04~m spacing, lies
under each sole; downward rays measure the signed deviation
$d_{f,i}$ of the terrain from the sole's nominal support plane,
positive above the plane, and a missed ray counts as unsupported. A
sample with $d_{f,i} < -0.11$~m overhangs the support, one with
$d_{f,i} > 0.02$~m penetrates a protrusion, and both are violations;
the stance feet accrue
\begin{equation}
r_t^{\mathrm{sole}} = -\lambda_s \sum_{f \in
\mathcal{F}_{\mathrm{st}}} \frac{1}{M} \sum_{i=1}^{M}
\mathbf{1}\bigl[\, d_{f,i} \notin [-0.11,\, 0.02] \,\bigr],
\label{eq:sole}
\end{equation}
where $\mathcal{F}_{\mathrm{st}}$ is the set of stance feet;
$\lambda_s = 0.5$, doubled on gaps and stairs. A foot near a step
or stone edge gains overhanging
samples immediately; the penalty activates and pushes the
foothold toward the support center.

\textbf{Slope-aligned velocity decomposition.} On slopes and
stairs, motion follows the terrain surface, so world-horizontal
velocity tracking does not match the actual motion on ascent and
descent. We fit a
least-squares plane to a privileged height scan of about $1.1 \times
1.1$~m centered on the base and project the horizontal command onto
the plane's tangent basis (Fig.~\ref{fig:framework}(h)),
\begin{equation}
\boldsymbol{v}^{*} = v_x^{\mathrm{cmd}}\, \boldsymbol{t}_x +
v_y^{\mathrm{cmd}}\, \boldsymbol{t}_y,
\label{eq:slope}
\end{equation}
where $\boldsymbol{n}{=}(a, 0, c) = (\sin\theta, 0, -\cos\theta)$ is
the unit downward normal at slope pitch $\theta$, $\boldsymbol{t}_x$ is
the fore-aft tangent, and
$\boldsymbol{t}_y = \boldsymbol{t}_x \times \boldsymbol{n}$;
velocity tracking follows $\boldsymbol{v}^{*}$. The
posture term keeps the base upright in the world rather than
aligned with the slope normal, decoupling velocity from posture; on
steep slopes this offsets the center of mass relative to the
support surface, and real-world tests validate slopes to
20$^\circ$ against the 31$^\circ$ curriculum maximum.

\textbf{Symbols.} Tracking errors are evaluated at the base; tilde
quantities are joint torque and velocity,
normalized by their limits, and the two actuator rows carry no
separate weight. Kernel bandwidths carry their arguments' units:
$\sigma_v{=}0.63$~m/s, $\sigma_\omega{=}0.63$~rad/s, $\sigma_g{=}1.0$
on normalized gravity, $\sigma_h{=}0.45$~m;
$v_{\mathrm{co}}{=}\max(\|\boldsymbol{v}^{\mathrm{cmd}}\|,1)$:
velocity-error scale; $h^{*}{=}0.885$~m: base-height target;
$\boldsymbol{e}_{t,xy}$, $e_z$, $\omega^{*}$: horizontal and
vertical command errors, yaw-rate command; $\omega_z$, $h$,
$\boldsymbol{v}_f$: base yaw rate, base height, stance-foot
velocity; $a_f$: stance-foot acceleration in m/s$^2$;
$(\cdot)^{+}$: positive part; $\mathrm{viol}_i$: joint $i$ beyond a
position, velocity, or torque limit; $\boldsymbol{g}$,
$\hat{\boldsymbol{g}}$: projected gravity and its upright reference,
both in the base frame;
$\mathcal{B}_{\mathrm{ill}}$: base, torso, hips, knees, shoulders, and
head.

\subsection{Unified Training with Deployment-Grade Robustness}
\label{sec:training}

\textbf{Multi-terrain curriculum.} Training runs 4096 parallel
environments in Isaac Lab~\cite{mittal2025isaaclab} on two RTX 5090s. Terrains form a
type-by-difficulty grid of ten uniformly spaced levels per type
(Table~\ref{tab:terrain}); each environment walks its terrain
track, difficulty following the
standard curriculum of massively parallel reinforcement
learning~\cite{rudin2021learning}: an environment passing the track
midpoint moves up a level, one covering less than half the commanded
distance moves down, and top-level environments fall back to a
random level. Episodes end when the base
sinks below the least-squares fit of a privileged height scan
of about $1.7 \times 1.1$~m centered on the base, or when illegal contact persists for
about 1~s.

\begin{table}[!t]
\caption{Difficulty curriculum ranges of the eight terrain types}
\label{tab:terrain}
\centering
\scriptsize
\setlength{\tabcolsep}{3pt}
\begin{tabular}{@{}lll@{}}
\toprule
Terrain & Difficulty parameter & Range, easy to hard\\
\midrule
Stairs & step height & 0.10--0.30 m\\
High platform & platform height & 0.25--0.85 m\\
Gap & gap width & 0.20--1.20 m\\
Slope & incline, rise over run & 0.20--0.60, about 11--31$^\circ$\\
Discrete obstacles & obstacle height & 0.10--0.23 m\\
Balance beam & beam width & 0.50$\to$0.30 m\\
Stepping stones & size / spacing & 0.60$\to$0.35 / 0.10$\to$0.50 m\\
Thin barrier & barrier height & 0.25--0.60 m\\
\bottomrule
\end{tabular}
\end{table}

\begin{table}[!t]
\caption{Training and method hyperparameters}\label{tab:hyper}
\centering
\scriptsize
\renewcommand{\arraystretch}{0.95}
\setlength{\tabcolsep}{2pt}
\begin{tabular}{@{}ll@{}}
\toprule
Parameter & Value\\
\midrule
\multicolumn{2}{@{}l}{\textit{PPO training}}\\
Optimizer & Adam ($\beta_1{=}0.9$, $\beta_2{=}0.999$)\\
Learning rate & $3{\times}10^{-4}$\\
Discount $\gamma$ / advantage $\lambda$ & 0.99 / 0.95\\
Clip range / gradient-norm cap & 0.2 / 1.0\\
Entropy / value-loss coefficient & 0.005 / 1.0\\
Steps per env. / minibatches / epochs & 24 / 4 / 5\\
\midrule
\multicolumn{2}{@{}l}{\textit{Method and robustness}}\\
Action scale / log-std clip & 0.25 / $[-6,2]$\\
Cloud noise $\sigma$ / dropout / outliers & 0.02~m / 4\% each\\
\bottomrule
\end{tabular}
\end{table}

\textbf{Perception-degradation injection.} The point pipeline
degrades throughout training (Fig.~\ref{fig:framework}(f); magnitudes
in Table~\ref{tab:hyper}): Gaussian noise before voxelization, point
dropout, outliers, a 20-Hz rotational-sweep LiDAR model, a random
0--1-frame buffer lag, and self-occlusion that drops rays hitting
the robot's own mesh. Real clouds still contain unmodeled swinging-limb returns
outside the static box; the learned attention
assigns them near-minimal weight (Section~\ref{sec:real}).

\textbf{Domain randomization.}
Actions execute after a per-environment random delay of 0--40~ms.
Observations receive uniform noise ($\pm0.07$ angular velocity,
$\pm0.035$ projected gravity, $\pm0.01$ joint positions, $\pm0.1$
joint velocities; observation units) and a random 0--1-substep delay.
Physics domain randomization covers friction 0.3--1.1, link masses
$\times$0.9--1.1, payloads, center-of-mass offsets,
PD gains $\times$0.8--1.2, motor strength, torque-speed saturation,
and voltage-dependent speed limits. Quantities unavailable at
deployment enter only the critic and the reward, so the policy's
observations match deployment exactly and transfer needs no
fine-tuning. Training takes about 40~h with standard PPO
settings~\cite{schulman2017ppo}, summarized in Table~\ref{tab:hyper}.

\section{Experiments}
\label{sec:exp}

\subsection{Experimental Setup}
\label{sec:setup}

\begin{figure}[!t]
\centering
\includegraphics[width=\columnwidth]{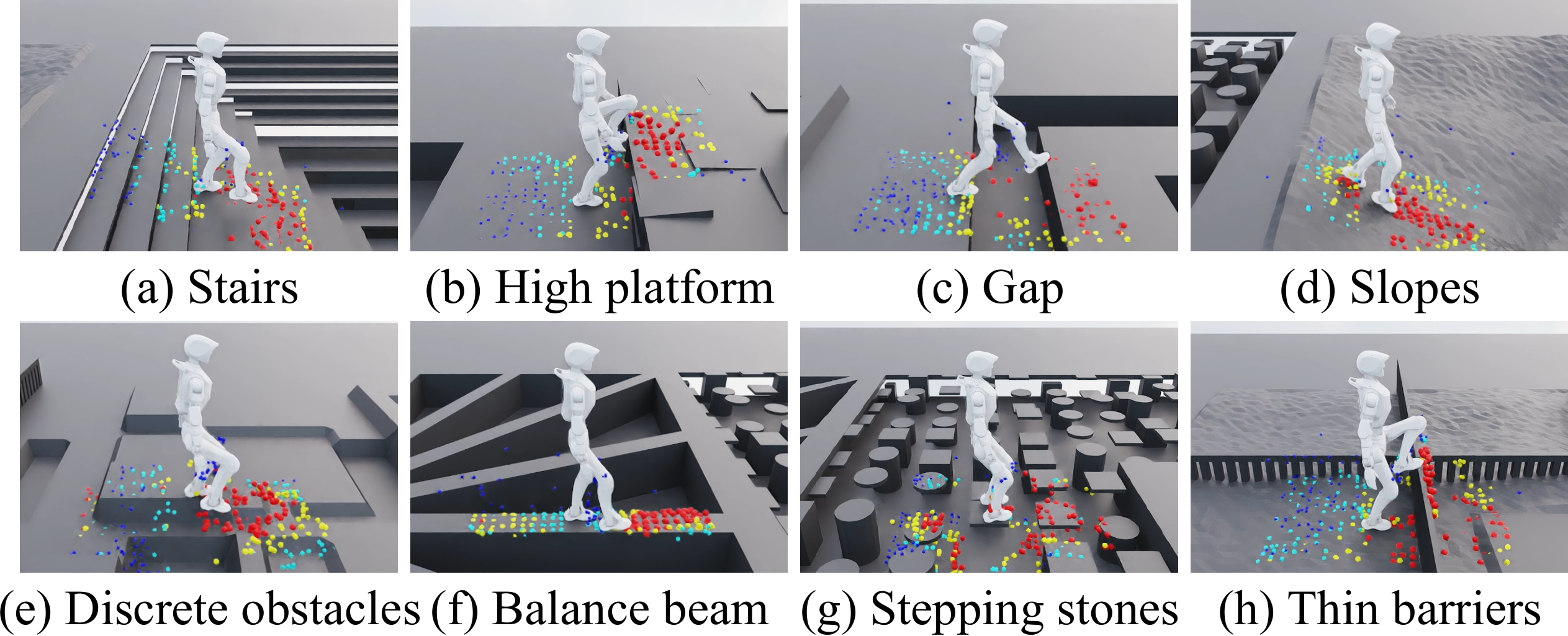}
\caption{Simulation snapshots of the eight terrain types of
Table~\ref{tab:terrain}; overlaid points are the voxel tokens
the policy reads, colored by online cross-attention weight, warm
is high, cold is low.}
\label{fig:simterrains}
\end{figure}

\begin{table}[!t]
\caption{Compute cost of the perception representations}\label{tab:compute}
\centering
\scriptsize
\setlength{\tabcolsep}{4pt}
\begin{tabular}{lccc}
\toprule
Representation & Params (M) & FLOPs (M) & Latency (ms)\\
\midrule
UniPoint, 0.1-m voxels & 1.83 & 5.2 & 1.5$\pm$0.11\\
UniPoint, 0.05-m voxels & 1.83 & 40.3 & 7.9$\pm$0.89\\
UniPoint, standard attention & 1.83 & 28.2 & 9.7$\pm$2.6\\
Depth-image CNN~\cite{sun2026dpl} & 4.25 & 91.0 & 8.6$\pm$1.29\\
\bottomrule
\end{tabular}
\end{table}

\textbf{Platform and deployment.} The robot is a DR02 humanoid from
Deep Robotics, 1.75~m and about 75~kg with 31 degrees of freedom;
the policy commands the 23 joints of Section~\ref{sec:overview} at
50~Hz, locks the other eight at zero, and reads the sensors of
Section~\ref{sec:fusion}. On an onboard RK3588, the policy forward
pass averages 1.5~ms against 8.6~ms for the depth-image CNN
baseline, over 100 ONNX Runtime CPU runs at batch~1, preprocessing
excluded; point preprocessing adds 0.2--0.5~ms.
Table~\ref{tab:compute} compares the perception representations in
parameters, floating-point operations (FLOPs),
and measured latency, with $\pm$ giving the standard deviation over the
100 runs.

\textbf{Baselines and ablations.} Every baseline retrains under our
stack, curriculum, and budget. The
\emph{blind} baseline reads only the proprioceptive
history~\cite{long2024hybrid}; the \emph{height-sampling} baseline
reproduces the perception representation of PIM~\cite{long2025learning},
whose open-source implementation targets a different platform: the baseline feeds a
0.1-m-resolution height map in the base frame directly to
the policy with matched noise~\cite{lee2020quadrupedal,miki2022learning},
the deployment form of elevation-map methods~\cite{wang2025beamdojo};
the \emph{depth-image
CNN} encodes both depth images with the convolutional design of
DPL~\cite{sun2026dpl} and carries no LiDAR; the \emph{concat}
variant swaps cross-attention for a multilayer perceptron over the same point
input. Two ablations remove the foot-sole scan and
the slope-aligned velocity decomposition.

\textbf{Metrics.} Success means crossing within the
terrain's time limit; falls, judged by the training termination
criterion (Section~\ref{sec:training}), and timeouts count as
failures; real-world
settings report successes out of 20 trials. The thin barrier is
rigid in simulation and, for safety, a freestanding plate the robot
can push over in real-world tests; knocking it over counts as a
failure. Each simulation cell
averages over 5 training seeds across 100
environments and 10 rollouts each, $n{=}5000$. The supplementary video shows
uncut runs of every terrain, failures included. Indoor
hazardous-terrain tests use a slack
overhead tether that bears load only in a fall; outdoor tests run
untethered.

\subsection{Simulation Results}
\label{sec:sim}

\begin{table}[!t]
\caption{Simulation success rates (\%)}\label{tab:sim}
\centering
\scriptsize
\renewcommand{\arraystretch}{1.05}
\setlength{\tabcolsep}{1.4pt}
\begin{tabular}{lcccccccc}
\toprule
Method & Stairs & Platform & Gap & Slope & Discrete & Beam & Stones & Barrier\\
\midrule
\textbf{UniPoint} & 99.5 & 98.6 & \textbf{99.6} & 83.5 & 99.6 & \textbf{97.2} & \textbf{79.2} & \textbf{99.8}\\
Blind & 78.9 & 12.1 & 21.3 & 81.8 & 93.6 & 9.6 & 8.3 & 6.7\\
Height sampling & \textbf{99.6} & 92.2 & 94.5 & \textbf{87.1} & 98.6 & 94.2 & 67.1 & 17.8\\
Depth-image CNN & 97.2 & 85.7 & 9.2 & 82.7 & \textbf{99.7} & 96.3 & 71.8 & 84.2\\
Concat & 98.5 & \textbf{98.9} & 98.1 & 84.4 & 99.6 & 95.7 & 69.3 & 89.5\\
\midrule
w/o foot-sole scan & 92.4 & 97.4 & 98.7 & 84.2 & 99.1 & 94.8 & 67.8 & 99.3\\
w/o velocity decomposition & 93.7 & 97.6 & 98.8 & 76.3 & 98.9 & 97.4 & 78.6 & 99.1\\
w/ standard attention & 98.7 & 98.2 & 99.4 & 85.1 & 99.8 & 96.6 & 80.9 & 99.6\\
w/ 0.05-m voxels & 99.1 & 96.9 & 99.8 & 84.7 & 99.3 & 95.8 & 82.3 & 99.4\\
w/ 0.2-m voxels & 94.2 & 95.1 & 95.8 & 81.4 & 96.9 & 90.6 & 67.9 & 91.2\\
\bottomrule
\end{tabular}
\end{table} 

Table~\ref{tab:sim} lists success rates in percent at the hardest
levels of all eight terrains (Fig.~\ref{fig:simterrains}); bold
marks the per-column best of the five compared methods, and the
ablation and variant rows below the rule do not compete. UniPoint
averages 94.6\% over the eight terrains and keeps every terrain at
79\% or above; each compared method falls below 70\% on at least
one terrain.
Blind
locomotion fails exactly where footholds must be seen: below 10\% on
thin barriers, balance beams, and stepping stones, with 12\% and
21\% on high platforms and gaps. Where the feet feel the terrain
for themselves---stairs, slopes, and discrete obstacles---it holds
79\%--94\%; footholds beyond the feet need exteroception.

Height sampling fails on vertical structures: 18\% on thin
barriers, because a top-down projection barely intersects the
standing plate, while the fused points keep its geometry
(Fig.~\ref{fig:simterrains}(h)) and UniPoint
reaches 99.8\%. Sparse footholds come next: 67\% on stepping stones
against UniPoint's 79\%, and 94\% against 97\% on the balance
beam. On stairs and slopes, which a complete height field
describes, it leads at 99.6\% and 87.1\%; the slope is the only
column where any baseline clearly leads UniPoint. Yet excluding
thin barriers, UniPoint totals 657 against height sampling's 633
over the other seven terrains.

The depth-image CNN has the lowest average of the perceptive
methods, 78\% against UniPoint's 95\%, at about $5.7\times$ the
inference latency (Table~\ref{tab:compute}): near parity on
stairs, slopes, discrete obstacles, and the balance beam, a
collapse to 9\% on gaps, and deficits of 7\% to 16\% on stepping
stones, high platforms, and thin barriers. It carries no LiDAR, so
part of the deficit comes from the sensor suite. The point-level
representation matches or beats it at under one-fifth of the
inference cost.

\begin{figure}[!t]
\centering
\includegraphics[width=\columnwidth]{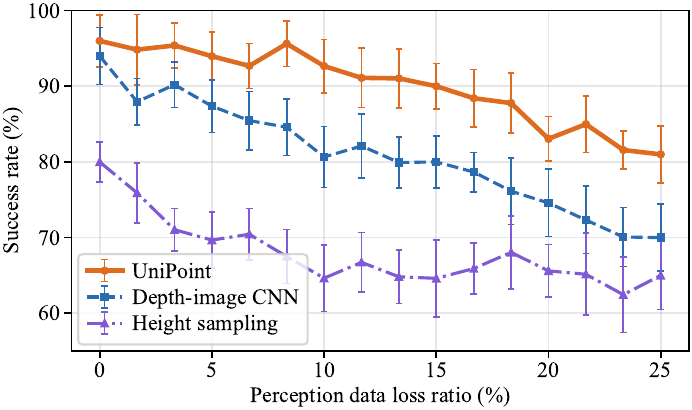}
\caption{Success rates under perception loss of increasing
severity, injected at test time. Error bars give the standard
deviation across training seeds; the vertical axis starts at
55\%.}
\label{fig:degradation}
\end{figure}

\begin{figure}[!t]
\centering
\includegraphics[width=\columnwidth]{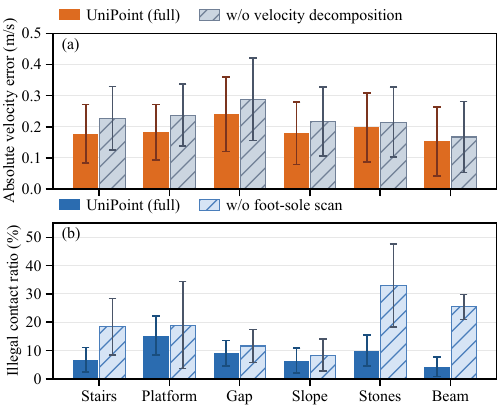}
\caption{Ablation metrics of the two terrain-aware rewards on six
terrain types. (a)~Absolute velocity error, the absolute deviation
of the magnitude of the base velocity from the 1-m/s command, for
the ablation without slope-aligned velocity decomposition.
(b)~Illegal-contact ratio for the ablation without the foot-sole
scan. Error bars give one standard deviation across
training seeds.}
\label{fig:ablation}
\end{figure}

\begin{figure}[!t]
\centering
\includegraphics[width=\columnwidth]{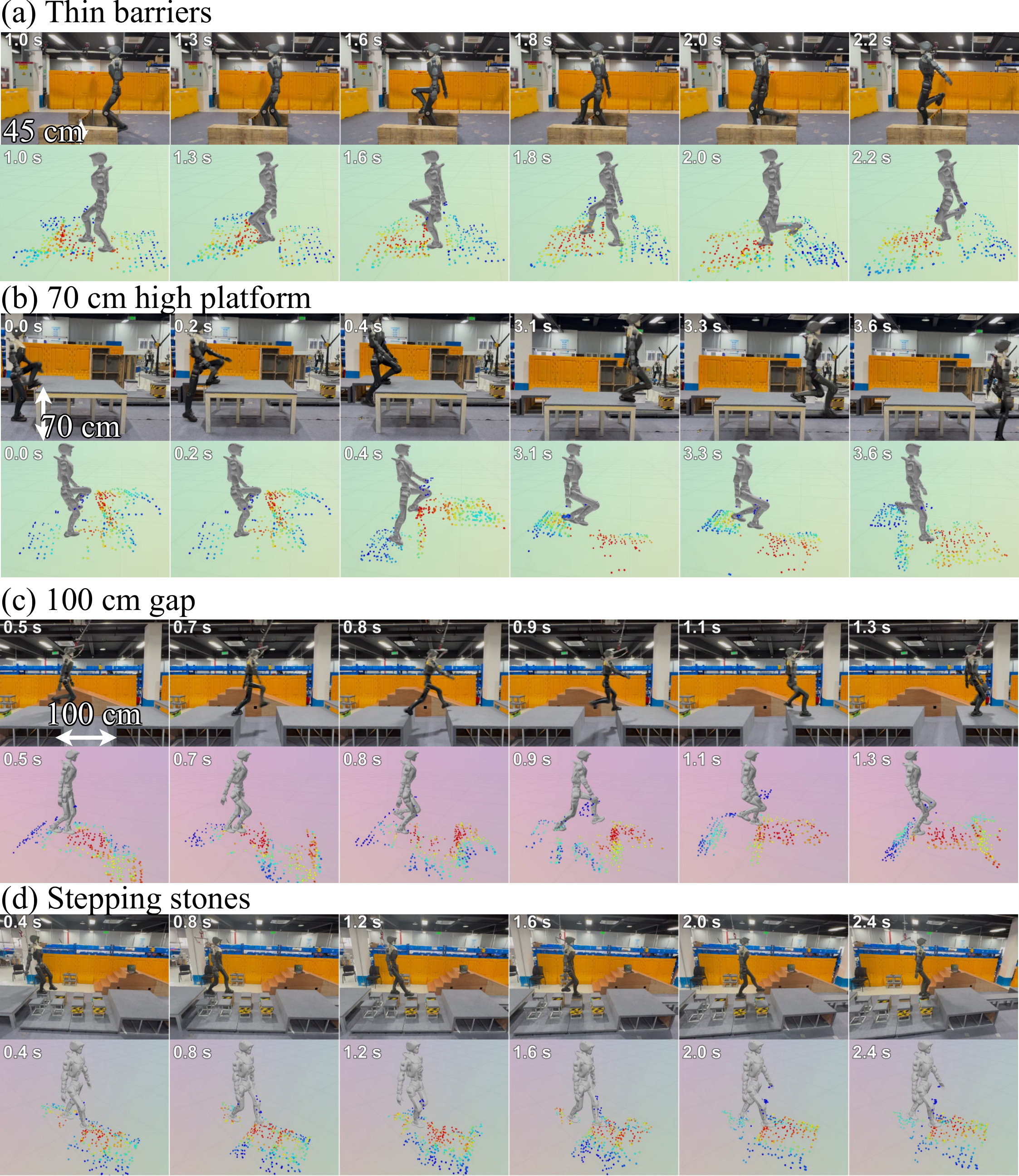}
\caption{Complete crossings at four representative settings of
Table~\ref{tab:real}. In each group, the top row shows robot
keyframes with time in seconds at the upper left, and the bottom
row shows the voxel tokens the policy reads at the same instants,
colored by online cross-attention weight; warm is high, cold is
low. In (b), the robot ascends and then descends the platform.}
\label{fig:experiment}
\end{figure}

Fusion quality rests on the proprioception-queried cross-attention:
replacing it with concatenation stays within 1.5\% on six terrains
yet drops stepping stones from 79.2\% to 69.3\% and thin barriers
from 99.8\% to 89.5\%---precisely the terrains whose decisive
tokens, stone
tops and plate surfaces, are a small minority of the input, and
isolating them needs queries that condition on state. Voxel size
sets a second tradeoff: the 0.05-m grid gains about 3\% on stepping
stones (82\% versus 79\%) at an unchanged average, loses 1\%--2\%
on high platforms and the balance beam, and costs about $5\times$
the latency (Table~\ref{tab:compute}); the unchanged coverage box
produces $8\times$ as many occupied voxels under the fixed
400-token budget. The 0.2-m grid drops across all eight terrains
by 2\% to 11\%, with the largest drops on stepping stones, thin
barriers, and the balance beam: coarse voxels push narrow supports
and thin
structures past resolvability. The 0.1-m grid is the balance
point.

\textbf{Perception degradation.} On stairs, balance beams, stepping
stones, and thin barriers, the terrains most dependent on
exteroception, we inject increasing test-time point loss with the
training dropout operator (Fig.~\ref{fig:degradation}). At about
4\%--5\% loss, UniPoint holds 94\%, the CNN 87\%, and height
sampling 70\%, each near its clean-input four-terrain level of
Table~\ref{tab:sim}; beyond that level, UniPoint keeps 81\% at
25\% loss and leads throughout, the CNN falling furthest, to 70\%,
and height sampling ending lowest, at 65\%. Shutting down one
entire modality still leaves UniPoint usable;
Section~\ref{sec:real} confirms this on the robot. Modality
redundancy in the fused point set converts directly into
degradation robustness.

\textbf{Ablations.} Removing the foot-sole scan drops stepping
stones from 79\% to 68\%, stairs from 99\% to 92\%, and the balance
beam from 97\% to 95\%, and raises the illegal-contact ratio from
10\% to 33\% on stepping stones and from 4\% to 25\% on the balance
beam (Fig.~\ref{fig:ablation}(b)): safe landing on sparse and narrow
supports rests on this dense signal. Removing slope-aligned
velocity decomposition drops slopes from 83\% to 76\% and stairs
from 99\% to 94\%; velocity
tracking oscillates during ascent and descent, and the absolute
velocity error at a commanded speed of 1~m/s rises on all six
terrains, from 0.19 to 0.23~m/s on average
(Fig.~\ref{fig:ablation}(a)). Swapping the linear layers
for standard softmax self-attention shifts the eight-terrain
average from 94.6\% to 94.8\%, every per-column change within
2\%: the linearization costs no measurable accuracy, and we
keep linear attention for the onboard headroom it leaves
(Table~\ref{tab:compute}).

\subsection{Real-World Results}
\label{sec:real}

\textbf{Quantitative traversal.} Fig.~\ref{fig:experiment} shows
complete crossings by one policy, with no terrain switch, reset, or
expert gating; Table~\ref{tab:real} reports successes out of 20
trials per setting, a dash marks settings a baseline cannot
complete, and a backward-stairs run appears below the rule. The nine
settings cover seven terrain types, with high platforms and gaps at
two difficulty levels each, and stepping stones pairing 35-cm tops
with 30-cm spacing. Most sit below the
training maximum: stairs and slopes at mid-level; platforms, gaps,
and barriers at 0.70, 1.00, and 0.45~m against 0.85, 1.20, and
0.60~m; the beam at its hardest; stone tops at the hardest size
(Table~\ref{tab:terrain}); success rates are therefore not comparable
across settings.

\begin{table}[!t]
\caption{Real-world quantitative results}\label{tab:real}
\centering
\scriptsize
\setlength{\tabcolsep}{2.2pt}
\begin{tabular}{lcccc}
\toprule
Terrain & Setting & UniPoint & Blind & Elevation map\\
\midrule
Stairs & 20-cm steps & 19 & 12 & 20\\
Slope & 20$^\circ$ & 20 & 16 & 20\\
High platform & 50-cm & 19 & --- & 19\\
High platform & 70-cm & 18 & --- & 9\\
Gap & 50-cm & 18 & --- & 16\\
Gap & 100-cm & 16 & --- & 4\\
Balance beam & 30-cm width & 19 & --- & 10\\
Stepping stones & 35-cm / 30-cm & 16 & --- & 2\\
Thin barrier & 45-cm height & 18 & --- & 0\\
\midrule
Backward stairs & same as stairs & 18 & 0 & ---\\
\bottomrule
\end{tabular}
\end{table}

\textbf{Real-world baselines.} The blind baseline retains only
stairs at 12/20 and slopes at 16/20. The elevation-map baseline deploys the
typical form, an online map built from a
single forward LiDAR and corrected by LiDAR-inertial
odometry~\cite{long2025learning,dong2025marg}; every other design
choice, reward, and curriculum step matches ours. It holds 20/20
on stairs and slopes, stays usable at the 50-cm levels, and
collapses to 9, 4, 2, and 0 of 20 on the 70-cm platform, 100-cm
gap, stepping stones, and thin barriers, with the balance beam at
10/20.

\begin{figure}[!t]
\centering
\includegraphics[width=0.99\columnwidth]{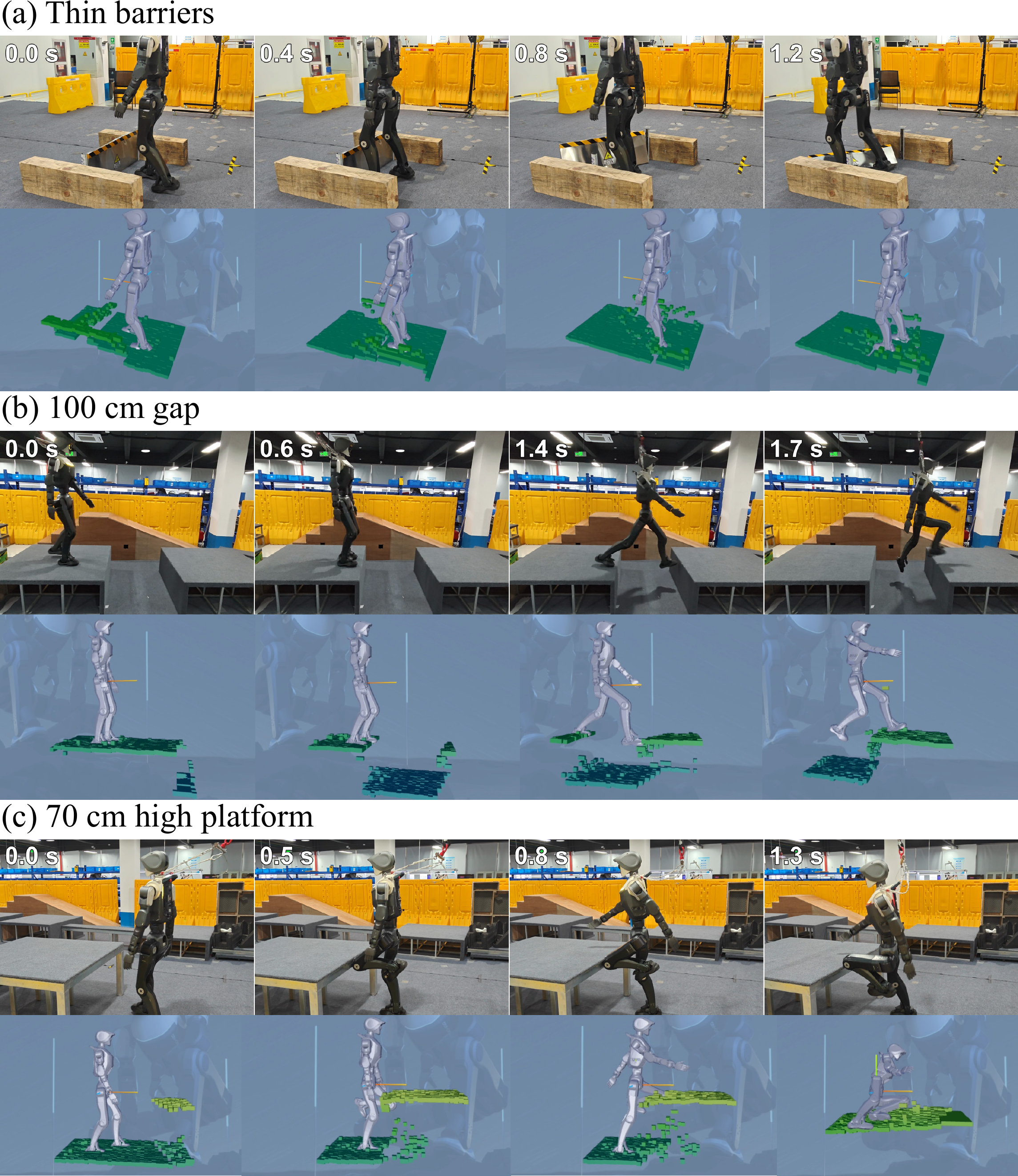}
\caption{Three representative real-world failures of the
elevation-map baseline of Table~\ref{tab:real}, which builds an
online elevation map from a single forward LiDAR. In each group,
the top row shows robot keyframes with time in seconds at the upper
left, and the bottom row shows the map the baseline builds at the
same instants.}
\label{fig:heightmap}
\end{figure}

\begin{figure}[!t]
\centering
\includegraphics[width=0.99\columnwidth]{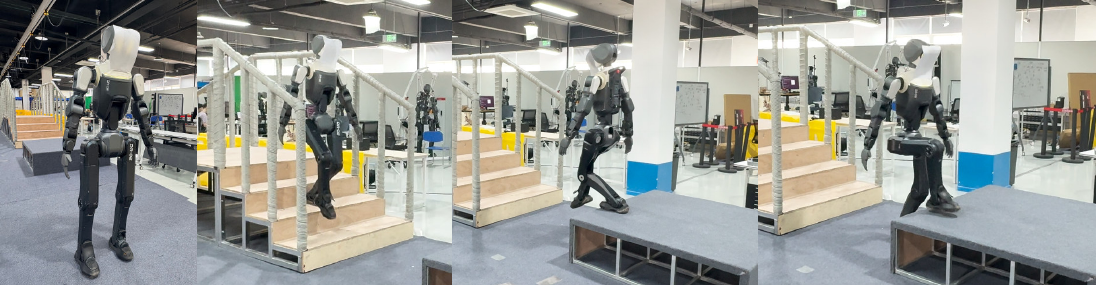}
\caption{Real-world occlusion test with the $360^{\circ}$ LiDAR
fully covered by a white bag; the leftmost panel shows the
occlusion setup. From left to right: descending the stairs,
stepping down from the high platform, and climbing onto the high
platform.}
\label{fig:occlusion}
\end{figure}

\textbf{Failure attribution.} This baseline changes both representation
and sensor suite, so its failures stem from that
deployment form; the representation-side control is the
height-sampling baseline, which reads privileged ground truth with
no mapping chain (Table~\ref{tab:sim}). The
standing thin barrier stays invisible to the map, matching the
simulated 18\%; the mapping chain itself breaks under aggressive
motion: fast pose change and robot vibration make the registration drift,
and sparse scans plus drift leave centimeter-level edge errors,
fatal for gaps, stepping stones, and balance
beams. Fig.~\ref{fig:heightmap} shows the three modes:
the barrier never enters the map and the robot knocks it over; the
far side of the 100-cm gap
maps as a thin strip, the front foot landing on its outer rim; the
70-cm platform breaks into scattered floating cells, and the swing leg
catches the lip. The forward-only LiDAR also sees nothing behind,
so the baseline cannot handle complex terrain backward; UniPoint,
keeping no global map and reading only local voxel tokens, walks
stairs backward at 18/20. Egocentric forward depth needs explicit
terrain memory for
this~\cite{luo2026lf2wb}, and multi-camera designs~\cite{zhang2026rpl}
and under-base reconstruction~\cite{song2025gait} support backward
walking with limited coverage.

Discrete obstacles stay simulation-only; we make no real-world
claim for them. Published platforms reach 45~cm~\cite{yu2026ssr},
and gaps reach 65~cm on a quadruped~\cite{dong2025marg} and 70~cm
to 1.2~m on humanoids~\cite{fu2026glad,yu2026ssr,li2026taga}, under
differing protocols. Our 70-cm platform exceeds the reported
heights, and our 100-cm gap sits in the upper part of the reported
range, all from one policy at about 1.5~ms onboard latency
(Table~\ref{tab:compute}); we target terrain breadth, policy
unification, and low-compute deployment over single-terrain
extremes. UniPoint's remaining failures concentrate
on unclear perception of terrain edges, leading to misplaced
footholds.

\textbf{Single-modality failure.} With either modality off, cameras
or LiDAR, the robot walks stably on flat ground and grass. With the
$360^{\circ}$ LiDAR fully covered (Fig.~\ref{fig:occlusion}) and only
the two depth cameras left, the robot climbs a 40-cm-high platform
and walks 20-cm stairs up and down. Both demonstrations stay at or
below the quantitative levels: traversal survives challenging terrain
under single-modality failure, the most direct real-world evidence
for point-level fusion.

\textbf{Attention behavior.} The online cross-attention weights, in
simulation (Fig.~\ref{fig:simterrains}) and in the real-world runs
of Fig.~\ref{fig:experiment}, show three consistent patterns: weight
concentrates on terrain the robot is about to use, before the
action; gap rims, platform lips, stone rims, and barrier tops
sustain the highest weight, the geometry that decides foothold
safety and takeoff timing; the robot's own points and regions
behind or away from the path sit near the minimum. These patterns
match the proprioception-queried cross-attention of
Section~\ref{sec:fusion}.

\begin{figure}[!t]
\centering
\includegraphics[width=0.99\columnwidth]{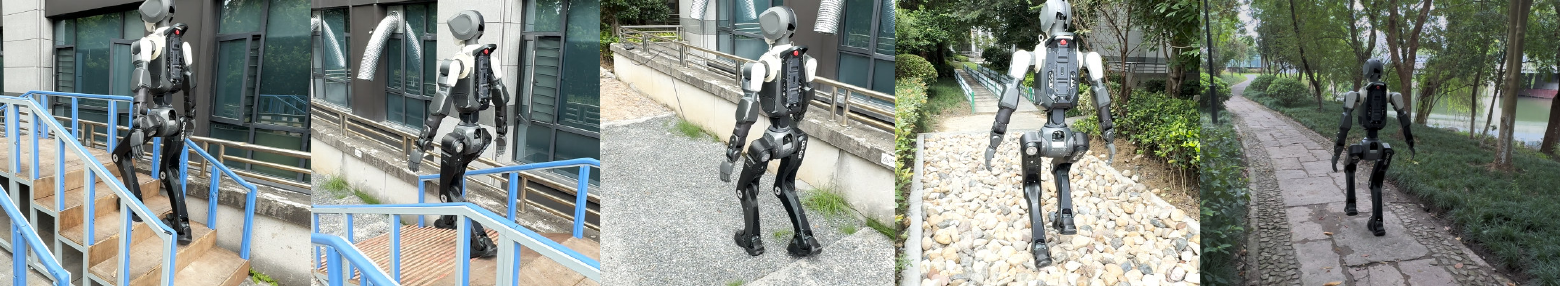}
\caption{Zero-shot generalization over outdoor unstructured
terrain, without fine-tuning or a tether. From left to right:
outdoor stairs, steel-grating stairs, gravel, a rocky path,
and a flagstone trail through forest.}
\label{fig:outdoor}
\end{figure}

\textbf{Outdoor generalization.} Training covered only the
parametric synthetic terrains of Table~\ref{tab:terrain}
(Fig.~\ref{fig:simterrains}), yet the
same policy deploys without fine-tuning or a tether onto the six
outdoor terrain types of Figs.~\ref{fig:teaser}
and~\ref{fig:outdoor} and walks stably:
point-level geometry learned in simulation transfers directly to
real outdoor terrain.

\section{Discussion and Conclusion}
\label{sec:conclusion}

UniPoint's effectiveness comes from three interlocking components. The
odometry-free point set retains standing thin barriers and survives
single-modality failure, with height sampling at 18\% against
UniPoint's 99.8\% in simulation. The foot-sole scan turns
the sparse fall signal into a dense one; removing it drops stepping
stones from 79\% to 68\% and stairs from 99\% to 92\%.
Perception-degradation injection and domain randomization carry the
policy onto the robot unchanged.

Sensing reaches about 1.3~m ahead and 1.1~m behind, a coverage
setting rather than a limit of the method; preprocessing grows
only linearly with point count, and the fixed token budget keeps
network compute constant. Point dropout and outliers partly cover
camera failures such as flying pixels and low-texture speckle, and
the real-world occlusion test (Fig.~\ref{fig:occlusion}), which
shuts down one entire modality, shows graceful degradation well
beyond these modeled failures.

A single training run across all eight terrain types produces one
policy. Quantitative real-world tests on the DR02 show that
point-level fusion achieves terrain breadth, policy unification, and
graceful degradation at constant forward cost. Future work will
extend the sensing range and refine the voxel size, add upper-body
contact and manipulation, and study perception fusion in dynamic
environments.

\bibliographystyle{IEEEtran}
\def\IEEEbibitemsep{0pt plus .5pt minus 1pt}

\bibliography{refs}

\end{document}